\documentclass[preprint,3p,twocolumn]{elsarticle}

\usepackage{lineno}
\modulolinenumbers[5]

\PassOptionsToPackage{hyphens}{url}\usepackage{hyperref}

\makeatletter
\def\ps@pprintTitle{%
   \let\@oddhead\@empty
   \let\@evenhead\@empty
   \let\@oddfoot\@empty
   \let\@evenfoot\@oddfoot
}
\makeatother

\usepackage{booktabs}

\begin{document}

\begin{frontmatter}

\title{The MICCAI Hackathon on reproducibility, diversity, and selection of papers at the MICCAI conference} 


\author[1,2]{Fabian Balsiger\corref{corr1}}
\ead{fabian.balsiger@insel.ch}
\address[1]{ARTORG Center for Biomedical Engineering Research, University of Bern, Bern, Switzerland}
\address[2]{Support Center for Advanced Neuroimaging (SCAN), Institute for Diagnostic and Interventional Neuroradiology, Inselspital, Bern University Hospital, University of Bern, Bern, Switzerland}

\author[1]{Alain Jungo\corref{corr1}}
\ead{alain.jungo@artorg.unibe.ch}

\author[3]{Naren Akash R J}
\address[3]{Center for Visual Information Technology, IIIT Hyderabad, Hyderabad, India}

\author[4]{Jianan Chen}
\address[4]{Department of Medical Biophysics, University of Toronto, Toronto, Canada}

\author[5]{Ivan Ezhov}
\address[5]{Technical University of Munich, Munich, Germany}

\author[6]{Shengnan Liu}
\address[6]{Department of Cardiology, Erasmus MC University Medical Center, Rotterdam, The Netherlands}

\author[7]{Jun Ma}
\address[7]{Department of Mathematics, Nanjing University of Science and Technology, Nanjing, China}

\author[5]{Johannes C. Paetzold}

\author[8]{Vishva Saravanan R}
\address[8]{IIIT Hyderabad, Hyderabad, India}

\author[5]{Anjany Sekuboyina}

\author[5]{Suprosanna Shit}

\author[1]{Yannick Suter}

\author[9]{Moshood Yekini}
\address[9]{African Masters of Machine Intelligence, Accra, Ghana}

\author[10]{Guodong Zeng}
\address[10]{sitem Center for Translational Medicine and Biomedical Entrepreneurship, Bern, Switzerland}

\author[11]{Markus Rempfler\corref{corr1}}
\ead{rmpflr@gmail.com}
\address[11]{Friedrich Miescher Institute for Biomedical Research (FMI), Basel, Switzerland}

\cortext[corr1]{Equal contribution and corresponding authors. The other authors are listed alphabetically.}

\begin{abstract}
The MICCAI conference has encountered tremendous growth over the last years in terms of the size of the community, as well as the number of contributions and their technical success. With this growth, however, come new challenges for the community. Methods are more difficult to reproduce and the ever-increasing number of paper submissions to the MICCAI conference poses new questions regarding the selection process and the diversity of topics. To exchange, discuss, and find novel and creative solutions to these challenges, a new format of a hackathon was initiated as a satellite event at the MICCAI 2020 conference: The MICCAI Hackathon. The first edition of the MICCAI Hackathon covered the topics reproducibility, diversity, and selection of MICCAI papers. In the manner of a small think-tank, participants collaborated to find solutions to these challenges. In this report, we summarize the insights from the MICCAI Hackathon into immediate and long-term measures to address these challenges. The proposed measures can be seen as starting points and guidelines for discussions and actions to possibly improve the MICCAI conference with regards to reproducibility, diversity, and selection of papers.
\end{abstract}


\end{frontmatter}


\section{Introduction}
The MICCAI conference is an annual scientific meeting for research in medical image computing (MIC) and computer assisted interventions (CAI). Approximately 2500~\citep{miccai-tweet2021} researchers attend the MICCAI conference every year to present and get to know the latest research related to MICCAI. With an acceptance rate of approximately 30~\%~\citep{Martel2020}, the MICCAI conference is highly competitive and can be considered among the top scientific conferences research in MIC and CAI. Besides the main conference that stretches over three consecutive days, there exist two satellite event days, one before and one after the three days, which allow the community to organize workshops, challenges, and tutorials dedicated to specific topics. At MICCAI 2020, for the first time, a hackathon was organized as such a satellite event.

The MICCAI Hackathon was initiated to benefit from the exchange among researchers during the MICCAI conference. A hackathon is especially well-suited as a satellite event at a conference, where junior and senior researchers go to exchange, discuss, and learn. Therefore, as a bottom-up approach, the MICCAI Hackathon benefits from this fruitful environment in order to foster collaborative work. The MICCAI Hackathon can be considered as a small think-tank rather than traditional workshops with talks and poster session, podium discussions, lecture-based tutorials, hands-on sessions, and challenges known from satellite event days. Accordingly, in this new format, participants gather and receive input from keynote speakers and mentors providing impulses about the hackathon's topic. The participants then work individually or together in teams to find solutions. Finally, they present their outcome at the end of the hackathon.

\begin{figure}
\centering
\includegraphics[width=.4\textwidth]{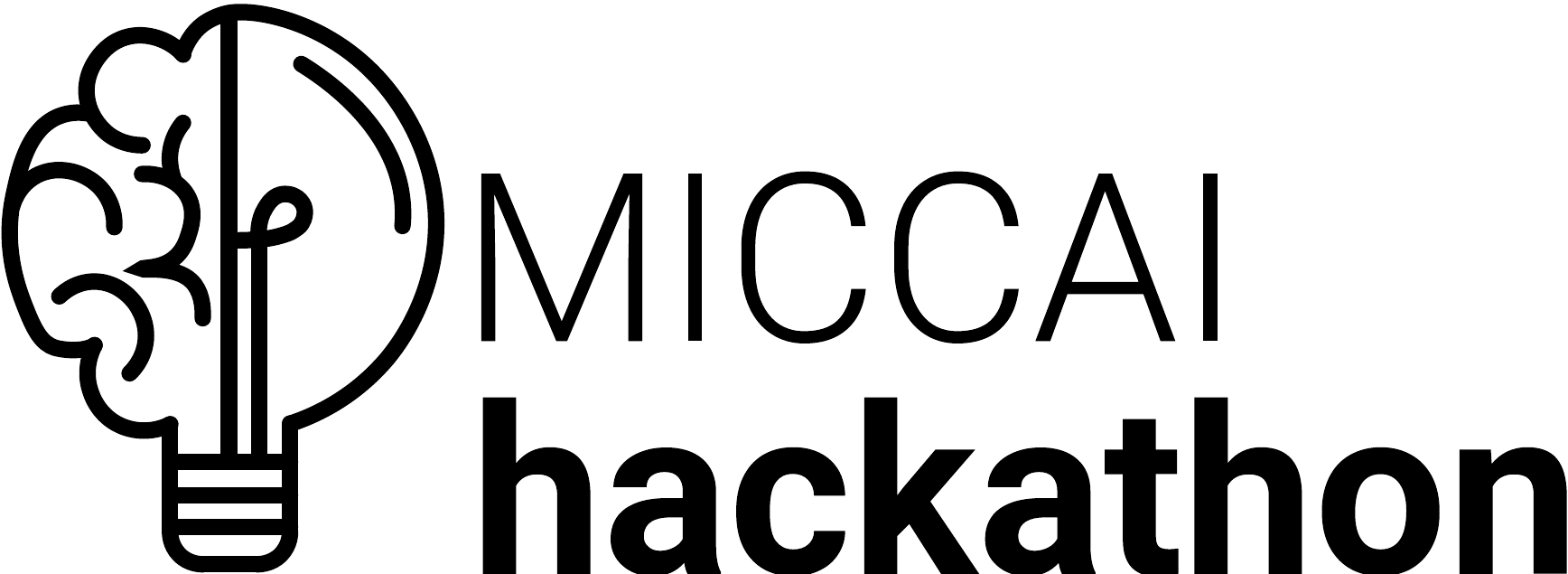}
\caption{The logo of the MICCAI Hackathon.}
\label{fig:logo}
\end{figure}

The first edition of the MICCAI Hackathon\footnote{\url{https://2020.miccai-hackathon.com/}} covered the topics reproducibility, diversity, and selection of MICCAI papers. The advance of machine learning has had a considerable impact on MICCAI, pushing the limits of algorithms, opening up completely new applications and ultimately, leading to an increased overall interest in MICCAI. With this success, however, came new challenges for the community. Complex, data-driven algorithms are more difficult to reproduce and the ever-increasing number of paper submissions to the MICCAI conference poses new questions regarding the selection process and the diversity of topics. To provide inputs to the participants regarding these topics, two keynotes were given and six mentoring sessions were held. Overall, five contributions were received providing ideas to tackle the challenges of reproducibility, diversity, and selection of MICCAI papers. For the interested reader, the contributions and keynotes of the MICCAI Hackathon are available online\footnote{\url{https://www.youtube.com/playlist?list=PLflMBx361ODCPE3TK-ROFSblPMK5cNOto}.}, and the procedure is detailed in the supplementary material.

The outcome of the first MICCAI Hackathon is summarized in this article. For the three topics reproducibility, diversity, and selection, \emph{immediate} and \emph{long-term measures} are presented in Section~\ref{sec:reproducibility} to~\ref{sec:selection}. \emph{Immediate measures} might be realizable for upcoming MICCAI conferences and \emph{long-term measures} would most likely require more time and effort to implement. All presented measures base on the keynotes, mentoring sessions, contributions, and opinions of the MICCAI Hackathon organizers. Finally, a short discussion concludes the article. We hope that the distilled measures could help to potentially improve MICCAI, as reproducibility, diversity, and selection of MICCAI papers concern the whole MICCAI community.

\section{Reproducibility}
\label{sec:reproducibility}
Reproducibility refers to that results of an experiment should be achieved with similar or equal results when the experiment is performed by another researcher. In MICCAI, this often involves executing computational methods on medical image data. However, this is not trivial to achieve as most methods are highly specialized and often not publicly available. For instance, the collection of MICCAI 2020 papers with code~\citep{miccai-open-source-papers} shows that there is still only a fraction of the papers that share their code and thereby facilitate reproducibility. Further, as MICCAI often involves protected medical data, sharing that data might not be straightforward or even impossible. Therefore, in the category reproducibility, two questions were specifically investigated:

\begin{itemize}
    \item What does it need for a MICCAI paper to be reproducible?
    \item What could MICCAI do to encourage reproducibility?
\end{itemize}

Three out of five contributions addressed the category of reproducibility. Overall, reproducibility was the aspect with the most frequent feedback that an improvement is desired. It is likely the category where measures are the most straightforward to implement and where we consider the potential for immediate improvements to be the highest. This is mostly due to the fact that MICCAI can benefit from the experiences of measures taken at other conferences (e.g., NeurIPS~\citep{pineau-2020-report}).

\subsection{Immediate Measures}

\textbf{Incorporate the reproducibility checklist in the paper submission form (at intent-to-submit \& final submission).} Incorporating a reproducibility checklist to the paper submission might raise awareness to this important topic and nudge authors towards adding important details to their manuscript. This measure has been effective at NeurIPS 2019~\citep{pineau-2020-report}. The initial checklist can adhere to the list by~\cite{pineau-2020-checklist} without adaptions. We consider it important that the list is already present at the intention to submit, but with the possibility to update the choices at the time of the final submission. By having such a list in the submission form, papers might be improved upon submission regarding reproducibility as people tend to forget, not intentionally, certain aspects when writing their papers (e.g., software used, annotation protocol, pre- or post-processing steps, \ldots). Besides, priming the authors at the intent-to-submit with the reproducibility checklist may create a so-called "mere-measurement effect" known in behavioural economy (see \cite{greenwald-1987,morwitz-2004,thaler-2009}) leading to more reproducible manuscripts. Feedback of the authors regarding the reproducibility checklist is essential. A questionnaire could assess the acceptance of it in the community, but likely requires determining a reproducibility chair at MICCAI.

\textbf{Introduce a reproducibility chair at MICCAI.} Having a reproducibility chair demonstrates that reproducibility is taken seriously and gives credit to the individual in charge of it. This has been shown to be effective at NeurIPS, but also examples like the MICCAI challenge chair indicate that this would have a positive effect. Amongst the chairs' responsibilities would be to analyze the influence of a reproducibility checklist over the years, gather feedback from authors and reviewers, and adapt the checklist (as well as potential other measures) to the domain-specific needs of MICCAI.

\textbf{Include a statement of data availability in the conference management toolkit (CMT).} Including a separate statement in the CMT (e.g., 250 words) on the availability of data could raise awareness to release data. The statement could address questions such as \textit{Is the used data publicly available, if not, why? Are comparable synthetic datasets available?}

\textbf{Promote reproducibility efforts of authors.} The visibility of authors with increased efforts regarding reproducibility should be enhanced. One way to achieve this could be an official list of open source MICCAI papers which could be either on the MICCAI society website and/or even be incorporated into the proceedings.

\textbf{Communicate best practices on reproducibility and code submission.} Similar to the guidelines for authors and reviewers, there could be guidelines for reproducibility (or a dedicated section in the guidelines for authors) pointing to the reproducibility checklist~\citep{pineau-2020-checklist} as well as resources like a code completeness checklist~\citep{code-completeness-checklist}. Similar efforts at other conferences have already been made (e.g. at the EMNLP conference~\citep{emnlp-post,emnlp-call-papers}) and do not need to be recreated from scratch.

\subsection{Long-term Measures}
\textbf{Introduce a reproducibility award.} A \textit{reproducibility award} might promote efforts towards more reproducible papers. The award could consider the reproducibility checklist, and papers with public code could be (ideally automatically) compared to the ML code completeness checklist on a specific date prior to the MICCAI conference, e.g., the camera-ready submission deadline. Specification of the exact procedure and forming of the committee could become one of the responsibilities of the MICCAI reproducibility chair.

\textbf{Code submission.} Submitting code should be considered as part of the paper submission (e.g., in form of a ZIP archive). Tools for code anonymization exist (e.g., search and replace for certain keywords) and the submitted code would not be released publicly. Such a code submission might further raise awareness and provide the reviewers with an additional way to assess papers (e.g., in case of doubt regarding an implementation). The code submission could even be delayed until the paper matching has completed.

\textbf{Best practices for evaluation.} Best practices for evaluation could be released by the MICCAI society. Separated into broad tasks (e.g., classification, segmentation, reconstruction), lists of metrics and their reference implementations could be provided. Further, guidelines for cross-validation, ablation, baselines, etc. could help authors to improve their papers without imposing extra effort and would highly increase comparability across papers. We see that efforts to provide reference implementations of metrics are already going on for MICCAI challenges and believe that these could most likely be embraced by the society once available.

\section{Diversity}
\label{sec:diversity}
The MICCAI conference attracts approximately 2500 attendees from all over the world every year. With this outreach, however, come questions about the diversity of the attendees and the topics that are covered by the scientific program. The participants, therefore, investigated the question:

\begin{itemize}
    \item Is MICCAI diverse enough?
\end{itemize}

One out of five contributions covered to the category of diversity. The MICCAI conference already puts much effort to increase the diversity of the attendees with initiatives like Women in MICCAI, RSNA panel discussion, student travel awards, student participation awards, and alike. Often, these initiatives promote equity, diversity, and inclusion (EDI), and are appreciated by the MICCAI community. Also, the virtual setting of MICCAI 2020 likely improved the diversity of attendees, and the contemplated joint physical/virtual MICCAI 2021 further substantiate these efforts of MICCAI to more diversity. Considering paper topic diversity, it appears to be unclear if and how more diversity should be increased. In general, we believe it is good that currently topic diversity is incorporated on the level of oral presentations such that underrepresented topics have a dedicated oral session(e.g., an oral session on reconstruction although only a small fraction of the papers address reconstruction).

\subsection{Immediate Measures}
\textbf{Include a statement of clinical relevance in the CMT.} The clinical relevance of a MICCAI paper is often not entirely clear. Including a separate statement in the CMT (e.g., 250 words) could help to understand the clinical relevance of a MICCAI paper, and might help to promote it. Such a statement could even be integrated into the review process (cf. Section~\ref{sec:selection}).

\textbf{Promote clinicians.} Clinicians are a minority at the MICCAI conference but ultimately our project initiators, data providers, collaborators, and customers. A dedicated clinical session where invited clinicians present their daily routine and problems could promote clinical discussion at the MICCAI conference. The format could be a parallel session, dedicated clinical presentations in each session, or as a panel discussion similar to MICCAI 2020. Here, also aspects of diversity could easily be taken into account (i.e., different problems in different countries).

\subsection{Long-term Measures}
\textbf{Open access proceedings.} Open access to the MICCAI conference proceedings would further promote diversity, as people and laboratories with limited financial resources get free access to the latest research in MICCAI.

\textbf{Unconventional submission award.} An specific award for an unconventional submission might promote the submission of papers covering less established topics. Such papers might be at the periphery of MICCAI, connect rather disjoint topics or identify previously ignored, but clinically relevant problems. Determining the award candidates might even be a side product of the categorization of accepted papers for the poster sessions. Papers that are not trivial to cluster might be good candidates. Alternatively, the reviewers or area chairs could promote such exotic papers for this award, similar to the young scientist award.

\textbf{Encourage non-standard submissions in the call for papers.} The call for papers could include a statement that encourages submission of underrepresented topics and/or the analysis of methods that provide new insights rather than novel methodology.

\textbf{Clarify the importance of achieving state-of-the-art.} Challenges specifically exist to advance the new state-of-the-art regarding a certain task. Hence, contributions targeting challenge datasets that "merely" improve state-of-the-art might be better placed at the corresponding challenge. On the other hand, papers with interesting ideas that do not achieve state-of-the-art might be worth being discussed at the main conference. An explicit statement in the guidelines for authors and reviewers could clarify this circumstance.

\textbf{Paper format guidelines.} Further improvements to the recently loosened page limit might be possible, as often size (and readability) of figures is sacrificed due to paper length restrictions. Ideally, only the word count of the main text (without title, authors and affiliations, abstract, figure and table captions, acknowledgement, and references) is restricted. The amount of words can be chosen such that it meets approximately the current extent of MICCAI papers. The figures and tables could be limited in numbers, e.g., a total of five, but the captions itself should be without or weak restrictions in number of words. Moreover, the restriction of two pages references could be further weakened. Such adaptions contribute to diversity (e.g., more figures are possible to add) and also to reproducibility (e.g., more thorough descriptions of methods are possible).

\section{Selection}
\label{sec:selection}
The papers that are submitted to the MICCAI conference undergo a scientific peer-review process to select the papers being presented at the conference. This selection process is often subject to intensive discussion among the authors of papers, and might be perceived unfair in certain cases. Therefore, in the category selection, two questions were investigated:

\begin{itemize}
    \item What could MICCAI do to improve the review process?
    \item What defines a good MICCAI paper?
\end{itemize}

One out of five contributions was in the category of selection. It is likely the category where measures are the most difficult to take and require careful considerations and implementation. Despite this, there are certain points that would be fairly easy and fast to address.

\subsection{Immediate Measures}
\textbf{Open reviews.} The reviews should be visible to everyone. After acceptance of the papers, the reviews could be released on the MICCAI society website or in a public repository. This might foster less aggressively worded reviews. Further, accessible reviews might be useful for new members of the community, new reviewers, and also promote further research ideas. At least, the reviewer of a paper should see the other reviews after paper decision such that reviewers can learn from each other.

\textbf{Better acknowledgement of reviewers.} The MICCAI conference acknowledges the best reviewers every year in the closing session. However, the visibility of the best reviewers is limited to a very short window within all other topics in the closing session. Better acknowledging the best reviewers, e.g., permanently on the MICCAI society website or by bold-facing the reviewers in the proceedings would be desirable. Further, being more transparent about the selection of the best reviewers would be necessary, e.g., \textit{What defines a good MICCAI paper review?}.

\textbf{Stable paper scoring scale.} The scoring system of the review process was subject to heavy changes in the last years, converging to the NeurIPS scoring in the 2020 edition of MICCAI. The current scoring is clear and well-understandable, and the rating scale should be kept constant over the years. Currently, only slight modification regarding wording might be necessary. Large changes from year to year should be avoided to prevent confusion.

\subsection{Long-term Measures}
\textbf{Interactive rebuttal phase.} The rebuttal phase is an aspect of the review process with high controversy, and several pros and cons were raised during the MICCAI Hackathon. The rebuttal might possibly be improved when an interaction between authors and reviewers is allowed. Such interactions could avoid misunderstandings. Important is that a reviewer sees only his own review and is not influenced by the other reviewers. Also, It should be possible to revise the review score. Ideally, reviewers could have a look at revised versions of a submission but this might be difficult due to the tight timeline of the conferences. Also, public reviews (similar to OpenReview) should not influence the official review process.

\textbf{Adapt paper matching.} The paper-to-reviewer matching is an aspect, where improvement is certainly possible but might also be difficult due to software constraints (TPMS, CMT). Ideally, the matching would consider the seniority of reviewers, e.g., match a senior and two less experienced reviewers. Tracking reviewers over the years, see next point below, might be required to implement this. Further, domain conflicts (e.g., two reviewers from the same laboratory) need to be restricted. The unwillingness expressed by the reviewers in the bidding process should be automatically taken into account in the matching algorithm (e.g. as an extra set of constraints) and not require tweaking of the weights and manual verification as of today.

\textbf{Introduce reviewer tracking.} A tracking of the reviewers over years might improve the review process~\citep{neurips-review2020}. Senior reviewers could be identified and matched together with less experienced reviewers. Flagging reviewers (e.g. as \emph{fails to meet expectations}, \emph{meets expectations}, \emph{exceeds expectations}; see NeurIPS~\citep{neurips-review2020}) should be possible. Both seniority and flags (from previous years) could be taken into account by the area chairs when assigning papers.

\textbf{Discussion platform for accepted papers.} A public discussion on an official platform (similar to OpenReview) should be possible after the acceptance of papers. The platform should feature the official and anonymized reviews but also allow to ask questions in an official communication channel to the authors.

\textbf{Introduce clinical feedback.} Implementing a clinical feedback might improve aspects of clinical relevance. Such a clinical feedback needs to be additional to the three reviewers (sort of a \emph{fourth reviewer}). For instance, it could base on the clinical statement in the CMT, and maybe involve having a look at the results (figures and tables). This short feedback could be taken into account by the area chair in case of borderline decisions or conflicting reviews. At least, it would provide interesting and important insights on MICCAI papers from a clinical perspective. A challenge, however, might be to enthuse enough clinicians to participate in such a review. A first step might be a dedicated clinical session as proposed in Section~\ref{sec:diversity}, \textbf{promote clinicians}.

\section{Discussion}

The first edition of the MICCAI Hackathon covered the topics reproducibility, diversity, and selection of papers at the MICCAI conference. These topics are already known to the MICCAI community, and are often informally discussed among colleagues during the MICCAI conference. During the MICCAI Hackathon, researchers of the MICCAI community collaborated and proposed possible solutions to challenges regarding these topics. The outcome of this hackathon during the satellite events has been summarized in Section~\ref{sec:reproducibility} to \ref{sec:selection} into \emph{immediate} and \emph{long-term measures} that could potentially be implemented by future organizers of the MICCAI conference and the MICCAI society. Most of these measures are low-risk and might be applied with manageable effort. Indeed, after sharing the herein presented measures with the MICCAI board and the organizers of MICCAI 2021, a reproducibility checklist has already been implemented as part of the paper submission for MICCAI 2021\footnote{\url{https://miccai2021.org/en/CALL-FOR-PAPERS.html}}.

The proposed measures are opinions of the organizers of the MICCAI Hackathon and are based on their own experience, discussions with the keynote speakers and mentors, and the participant's outcomes. Therefore, the measures are subjective and do not represent the opinion of all stakeholders involved at MICCAI. Further thoughts, ideas, and also disagreement with the suggestions are likely present in the community. Therefore, it would be essential to form a working group involving different stakeholders (PhD students, postdocs, faculty, MICCAI chairs, MICCAI board members, clinicians, ...) to adapt and refine the proposed measures further. Here, a Delphi process~\citep{Dalkey1963} could help arrive at a consensus originating from a representative group of stakeholders. This white paper could serve as an inspiration and starting point for such a process.

Besides the discussed topics, we also gained insights about the format of a hackathon in the context of MICCAI. The collaborative nature of a hackathon introduces new possibilities to the MICCAI community and fits well in the setup of the MICCAI conference. The first edition of the MICCAI Hackathon showed that this format can be useful for exchange of ideas and finding creative solutions. Instead of competing, as known from challenges, the participants are required to collaborate and profit from the enriching environment at the MICCAI conference. This collaborative aspect was highly endorsed by the participants, and also the keynote speakers and mentors were open to this new format and very supportive. This enables community-driven impulses as, for example, showcased by the distilled measures. More information on the hackathon's organization and experiences with it can be found in the supplementary material.

In conclusion, the first edition of the MICCAI Hackathon shed some light into ways to improve the aspects of reproducibility, diversity, and selection of papers at the MICCAI conference. The proposed measures can be seen as starting points and guidelines for further discussions and actions to possibly improve the MICCAI conference with regards to reproducibility, diversity, and selection of papers.

\section*{Acknowledgement}
The MICCAI Hackathon was supported by the Fund for the Promotion of Young Researchers (Nachwuchsförderungs-Projektpools) of the Intermediate Staff Association (Mittelbauvereinigung) of the University of Bern, grant attributed to FB and AJ. The organizers are very grateful to Anne Martel and Koustuv Sinha for giving the keynotes. The organizers are further very grateful to Anne Martel for sharing the tabular data regarding the paper selection at MICCAI 2020. A special thank goes to Marleen de Bruijne, Mattias Heinrich, Georg Langs, İlkay Öksüz, Lena Maier-Hein, and Koustuv Sinha for the fruitful discussions during the mentoring sessions. The organizers further thank the members of the award committee, Wilson Silva, Denis Schenk, Raphael Meier, and Mauricio Reyes for judging the contributions.

\bibliographystyle{model2-names.bst}\biboptions{authoryear}
\bibliography{references}

\section*{Supplementary Materials}

\begin{figure*}
\centering
\includegraphics[width=.8\textwidth]{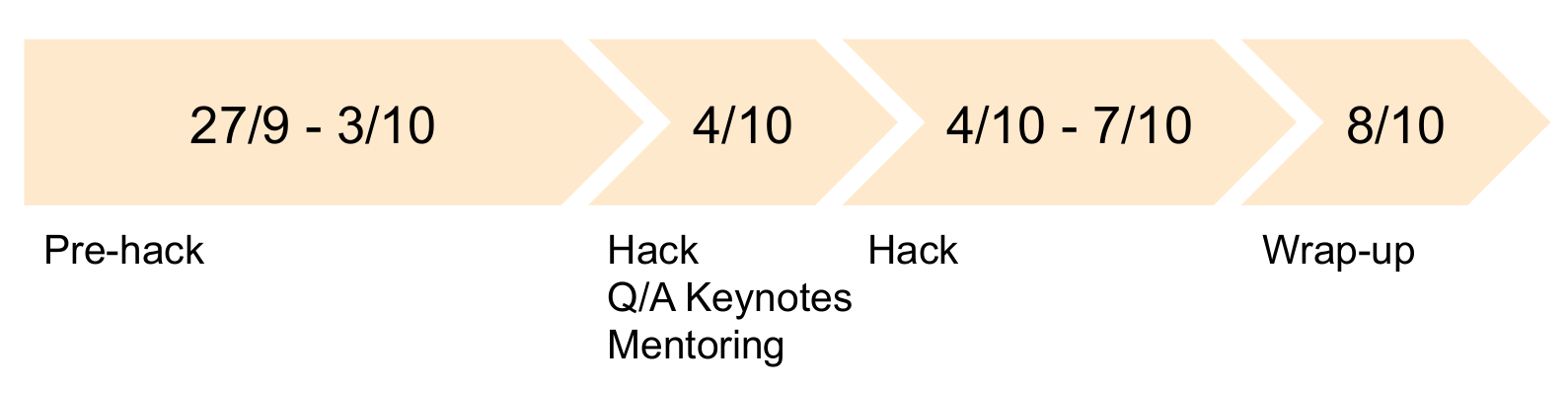}
\caption{A chronological overview of the MICCAI Hackathon in four phases. During the pre-hack phase, the keynotes and resources were made available to the participants. On the 4\textsuperscript{th} October, the actual event took place within the frame of the satellite event days. The keynote speakers and the mentors joined for live sessions scattered throughout the day to answer questions and interact with the participants. Then, the participants had time to work on their topic(s) and to provide their presentations until the 8\textsuperscript{th} October. On the 8\textsuperscript{th} October, the second satellite event day, the presentations were made available online and the participants joined for a public wrap-up session.}
\label{fig:overview}
\end{figure*}

The MICCAI Hackathon was organized by Fabian Balsiger, Alain Jungo, and Markus Rempfler. Being considered as part of the tutorial track by the MICCAI conference, the MICCAI hackathon proposal underwent review in this category (and faced some difficulties by \textit{not really} being a tutorial). Tutorials at the MICCAI satellite event days are either half-day or full-day events. For the hackathon, a full-day event seemed to be the minimum time required for the participants to work on their ideas, ideally, the available time should be even longer. As the COVID-19 pandemic led the MICCAI organizers to opt for a virtual rather than a physical conference, the initial organizational plans were slightly adapted because collaboration in a virtual setting with participants originating from different time zones is very challenging. Therefore, in the end, we stretched the hackathon over several days and four phases, see Figure~\ref{fig:overview}. During the pre-hack phase, the keynotes and resources were made available to the participants. On the 4\textsuperscript{th} October, the actual event took place within the frame of the satellite event days. The keynote speakers and the mentors joined for live sessions scattered throughout the day to answer questions and interact with the participants. Then, the participants had time to work on their topic(s) and to provide their presentations until the 8\textsuperscript{th} October. On the 8\textsuperscript{th} October, the second satellite event day, the presentations were made available online and the participants joined for a public wrap-up session.

To adapt to the virtual setting, the keynotes of Anne Martel and Koustuv Sinha were pre-recorded\footnote{The keynotes are available online at \url{https://www.youtube.com/playlist?list=PLflMBx361ODCPE3TK-ROFSblPMK5cNOto}.}. Anne Martel shared her view on organizing the MICCAI conference from a program chair's perspective including the aspects of selection and diversity of MICCAI papers. Koustuv Sinha talked about reproducibility in machine learning form the perspective of a reproducibility co-chair at the NeurIPS conference. Therefore, both keynotes covered the topic of the hackathon. On the 4\textsuperscript{th} October, both keynote speakers joined for a live questions and answers session. In addition, five mentors also joined throughout the day in 20~minutes time-slots to answer questions of the participants. The mentors were selected based on their experience regarding the hackathon's topic and with equity, diversity, and inclusion (EDI) criteria in mind. The mentors were Marleen de Bruijne, Mattias Heinrich, Georg Langs, İlkay Öksüz, and Lena Maier-Hein. Further, Koustuv Sinha joined for an additional mentoring session to the keynote questions and answers session.

A set of exemplary questions regarding the hackathon's topic was prepared to facilitate the work of the participants\footnote{Refer to \url{https://2020.miccai-hackathon.com/\#topics}.}. Resources were made available to provide further input to the participants\footnote{The resources are available online at \url{https://airtable.com/shrfRGA9FrHArvkNT}.}. These resources comprised scientific papers, links to blog posts and code repositories, and also tabular data from the paper selection process at MICCAI 2020. The latter included i) anonymized data on paper decisions and information on accepted papers, ii) anonymized data on reviewers, i.e., bids, quotas, subject areas, affinities, and assignments to submitted papers, and iii) anonymized data on area chairs, i.e., subject areas, relevance and Toronto paper matching system (TPMS) affinities to submitted papers. With these resources, the participants were further guided in their work, and also had the opportunity to analyze and explore real tabular data concerning paper selection at MICCAI 2020. In general, however, the participants were free to explore their own questions and search for resources within the frame of the hackathon's topics.

The participants were required to register for the MICCAI Hackathon. Registration was either possible as an individual participant, as a participant that would like to be matched into a team, and as a participant of an existing team. Therefore, the participants specified their preferred topic they would like to work on during the registration, which made a matching into teams based on preference possible. In total, 21 participants registered. Six were individual participants, three participants were matched into one team, and twelve participants registered as five teams. Most participants were PhD students and the ethnicities were fairly diverse. To facilitate the collaboration among the participants, a Discord server was set up that allowed video conferencing, screen sharing, and text messaging. In the end, five teams handed in a contribution. The overall impact of the MICCAI Hackathon goes likely beyond the registered participants. For example, on the MICCAI 2020 conference platform, approximately 80 attendees enlisted to the MICCAI Hackathon and might have watched the keynotes and contributions, which are still available online.

Several lessons were learned in this first edition of the MICCAI Hackathon. First, it is difficult to find the target audience for such a new initiative, especially in a virtual setting. There were no dedicated channels available to contact the MICCAI conference attendees directly to inform them about the initiative, leaving the organizers to rely on social media channels such as Twitter. This is expected to be less problematic for subsequent editions of the MICCAI Hackathon. Second, it seems that participants are more drawn to solution-driven tasks/questions, e.g., create a machine learning reproducibility checklist specific for MICCAI, than open questions, e.g., how to ensure reproducibility when the data cannot be shared. Third, the hackathon format does not entirely fit in the half-day and full-day setting of the satellite event days because it relies on collaborative work that cannot typically be performed during one satellite event day. Therefore, prolonging the hackathon over the entire conference duration was necessary. In an ideal case, the hackathon would not be restricted to the satellite event time slot(s) and could even be a separate category in the satellite events with a dedicated call for topics to be addressed.

\end{document}